\documentclass{article}

\usepackage[preprint]{corl_2026} 

\usepackage{microtype}
\usepackage{graphicx}
\usepackage{subcaption}
\usepackage{booktabs} 

\usepackage{hyperref}

\usepackage{amsmath}
\usepackage{amssymb}
\usepackage{mathtools}
\usepackage{amsthm}

\definecolor{undefined}{RGB}{0,0,0}
\definecolor{barrier}{RGB}{112,128,144}
\definecolor{bicycle}{RGB}{220,20,60}
\definecolor{bus}{RGB}{255,127,80}
\definecolor{car}{RGB}{255,158,0}
\definecolor{consveh}{RGB}{233,150,70}
\definecolor{motorcycle}{RGB}{255,61,99}
\definecolor{pedestrian}{RGB}{0,0,230}
\definecolor{trafficcone}{RGB}{47,79,79}
\definecolor{trailer}{RGB}{255,140,0}
\definecolor{truck}{RGB}{255,99,71}
\definecolor{drivesurface}{RGB}{0,207,191}
\definecolor{otherlane}{RGB}{175,0,75}
\definecolor{sidewalk}{RGB}{75,0,75}
\definecolor{terrain}{RGB}{112,180,60}
\definecolor{manmade}{RGB}{222,184,135}
\definecolor{vegetation}{RGB}{0,175,0}

\title{DualPathOcc: Dual-Resolution BEV Encoder for 3D Occupancy Prediction}

\author{
  Lihao Qiu$^{1}$ \enskip Jian Chen$^{2}$ \enskip Ruihao Wang$^{3}$ \enskip
  Ramu Gautam$^{1}$ \enskip Mei Yang$^{1}$ \enskip Yingtao Jiang$^{1}$ \\[2pt]
  $^{1}$Department of Electrical and Computer Engineering, University of Nevada, Las Vegas \\
  $^{2}$Institute of Logistics Science and Engineering, Shanghai Maritime University \\
  $^{3}$School of Software Engineering, Xi'an Jiaotong University \\[2pt]
  \texttt{\{qiul1, gautar1\}@unlv.nevada.edu} \quad
  \texttt{\{mei.yang, yingtao.jiang\}@unlv.edu} \\
  \texttt{daniel.chenjj@gmail.com} \quad \texttt{ruihaowang@stu.xjtu.edu.cn}
}

\begin{document}
\maketitle


\begin{abstract}
  Predicting 3D occupancy from multi-view images requires preserving geometric detail during 2D-to-3D lifting while reasoning over sparse, volumetric scene representations. We present DualPathOcc, a camera-based framework that combines a Spatial Enhancer for high-resolution feature aggregation before BEV compression, a SENet-augmented dual-path BEV encoder for local--global context modeling, and height-aware weighted cross-entropy for near-ground occupancy. The final model is optimized with occupancy supervision and no explicit depth loss. On single-frame Occ3D-nuScenes, DualPathOcc achieves 37.37 mIoU. We further analyze how surface-centered depth targets interact with volumetric occupancy learning.
\end{abstract}



\section{Introduction}
	
\label{sec:intro}
Image-based 3D occupancy prediction has recently gained significant attention in autonomous driving because it estimates dense voxel-wise geometry and semantics directly from camera observations \citep{tian2023occ3d,huang2023tri,wei2023surroundocc,li2023fb,yu2023flashocc,zhang2023occformer,hou2024fastocc,wang2024panoocc,ma2024cotr,kim2025protoocc}. Compared with camera-based 3D object detection \citep{huang2021bevdet,wang2022detr3d,li2022bevformer,liu2022petr,xie2022m,li2023bevdepth}, occupancy prediction represents free space, scene structure, and semantic content on a unified 3D grid.

Despite its promise, predicting 3D occupancy from 2D images remains challenging. First, images contain only surface-level observations, making full volumetric reconstruction inherently ill-posed. Second, BEV representations are highly sparse due to the dominance of empty space \citep{liu2024fully}. Third, the task exhibits strong geometric imbalance across object scales, from large vehicles to small pedestrians \citep{caesar2020nuscenes}. Finally, occupied voxels are heavily concentrated near the ground plane, further aggravating distribution imbalance \citep{tian2023occ3d}.

To address these challenges, DualPathOcc first aggregates lifted features at high spatial resolution before compressing them to the output BEV scale. A dual-path encoder then combines local geometric detail with broader scene context, while Squeeze-and-Excitation attention \citep{hu2018squeeze} recalibrates sparse channel-to-height features. Height-aware weighting further emphasizes near-ground voxels. This occupancy-specific coordination of spatial resolution, local--global encoding, channel attention, and vertical weighting forms the central design of DualPathOcc.

Our contributions are summarized as follows:\\
1. We present DualPathOcc, an occupancy-specific framework that enhances lifted features before BEV compression and processes them with complementary local and global paths.

2. We combine channel recalibration with height-aware occupancy weighting to strengthen sparse BEV features and near-ground scene structure.

3. DualPathOcc achieves 37.37 mIoU on single-frame Occ3D-nuScenes, 0.33 points above ProtoOcc, and provides a diagnostic analysis of surface-centered depth supervision for volumetric occupancy.

\section{Related Work}

\subsection{Vision-Based 3D Occupancy Prediction}
3D occupancy prediction is a dense 3D scene understanding task that requires richer geometric and semantic representations than BEV detection. Existing methods explore different strategies for lifting 2D features into 3D space. TPVFormer \citep{huang2023tri} reconstructs voxel representations from three orthogonal planes, while SurroundOcc \citep{wei2023surroundocc} leverages multi-scale voxel features for fine-grained reasoning. FB-Occ \citep{li2023fb} combines forward and backward projection features to improve consistency, and FlashOcc \citep{yu2023flashocc} encodes height information along the channel dimension for efficient occupancy prediction. Transformer-based approaches such as OccFormer \citep{zhang2023occformer} model both local and global context via dual-path architectures. FastOcc \citep{hou2024fastocc} uses lightweight 2D feature interpolation for voxel reconstruction, while PanoOcc \citep{wang2024panoocc} exploits panoramic multi-view inputs for globally consistent occupancy estimation.

Recent scene-completion methods incorporate additional geometric or temporal context. BRGScene \citep{li2024brgscene} couples stereo geometry with dense BEV volumes on SemanticKITTI. HTCL \citep{li2024htcl} learns cross-frame affinity and dynamic temporal refinement on SemanticKITTI and OpenOccupancy, while Hi-SOP \citep{li2024hisop} aligns depth-confidence-aware geometry and pose-guided temporal context on SemanticKITTI and NuScenes-Occupancy. DualPathOcc primarily studies single-frame Occ3D-nuScenes with LSS lifting and channel-to-height 2D BEV processing; its temporal extension uses one aligned history frame. Because these methods differ in input context, target benchmark, hardware, and profiling protocol, Section~\ref{sec:efficiency} reports a matched computational comparison with FlashOcc.

\subsection{Dual-Resolution and Multi-Branch Encoder Networks}
Balancing spatial detail and semantic context is a key challenge in dense prediction tasks. Existing methods address this via multi-resolution or multi-branch architectures. RefineNet \citep{lin2017refinenet} fuses multi-scale features through refinement blocks, while DeepLabV3+ \citep{chen2018encoder} uses atrous spatial pyramid pooling and an encoder--decoder design to capture multi-scale context. HRNet \citep{wang2020deep} maintains parallel high- and low-resolution streams with repeated feature exchange to preserve spatial precision. Multi-branch designs further improve representation learning. PIDNet \citep{xu2023pidnet} introduces detail, semantic, and context branches with hierarchical fusion, and Dual-Resolution Network \citep{hong2021deep} employs two parallel streams for spatial detail and semantic reasoning with periodic interaction.

\subsection{Channel Attention Mechanism}

Channel attention improves convolutional networks by adaptively reweighting channel-wise features \citep{wang2020eca}. The Squeeze-and-Excitation (SE) block \citep{hu2018squeeze} introduces global pooling and a bottleneck MLP to model inter-channel dependencies and has been widely adopted in vision tasks, including real-time segmentation frameworks such as BiSeNet \citep{yu2018bisenet}. Subsequent methods improve efficiency and flexibility. Variants such as ECA \citep{wang2020eca}, CBAM \citep{woo2018cbam}, and SKNet \citep{li2019selective} further improve efficiency or extend attention to the spatial and dynamic-kernel domains. 

\section{Methodology}
\subsection{Problem Formulation}
In this work, we aim to predict the 3D occupancy of surrounding scenes with multi-camera images $I=\{I^1, I^2, ...I^N\}$. Formally, the 3D occupancy prediction is represented as:
\begin{equation}
    V = G(I^1, I^2, ...I^N)
\end{equation}
where $G$ denotes the prediction network and $V \in \mathbb{R}^{H\times W\times Z}$ is the occupancy field on an ego-centric voxel grid. Extending the output to $L$ channels yields class-wise semantic occupancy probabilities $V_{\mathrm{sem}}\in\mathbb{R}^{L\times H\times W\times Z}$, where $L$ is the number of semantic classes.

\begin{figure*}[!t]
  \centering{\includegraphics[width=1\linewidth]{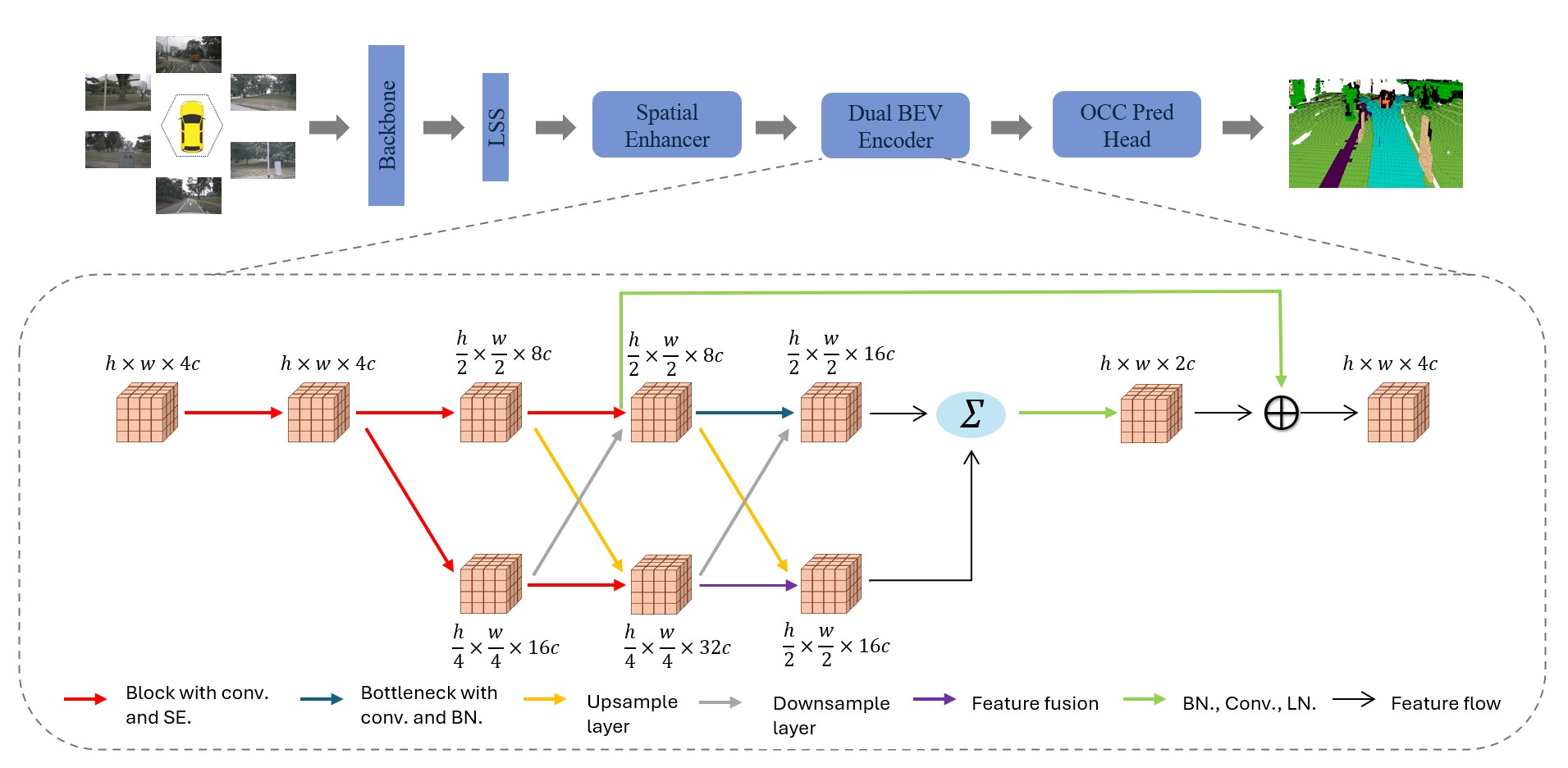}}\\
  \caption{Overview of DualPathOcc. Multi-view features are lifted by LSS, refined by the Spatial Enhancer before BEV compression, and processed by local and global encoder paths with channel recalibration before occupancy prediction. $\Sigma$ denotes element-wise addition and $\oplus$ denotes channel concatenation. The final model uses no explicit depth loss; depth-target variants are analyzed separately in Section~\ref{sec:gaussian_depth}.}
  \label{figure:network_structure}
\end{figure*}

\subsection{DualPathOcc}
Based on our observations, we introduce \textbf{DualPathOcc} (Figure~\ref{figure:network_structure}), a framework designed to address the challenges described in Sec.~\ref{sec:intro}. The architecture consists of four components: a 2D backbone for multi-view feature extraction, a Lift-Splat-Shoot (LSS) module for voxel-space projection, a spatial enhancement module for refining fine-grained BEV features, and a SENet-augmented dual-path BEV encoder for adaptive feature selection and geometric modeling. To further emphasize near-ground occupancy, we adopt a height-aware weighted cross-entropy loss that biases learning toward lower-elevation voxels.

\subsubsection{Image Encoder}
The image encoder comprises a backbone and a neck module. When processing multi-camera images at time $t$, the backbone extracts multi-level semantic features, which the neck module then aggregates to form comprehensive multi-level semantic information. We employ ResNet \citep{he2016deep} as the backbone architecture and FPN \citep{lin2017feature} as the neck module. The multi-level residual architecture in ResNet facilitates the extraction of rich semantic features across various levels, which FPN efficiently aggregates to produce high-quality 2D features.

\subsubsection{2D to 3D View Transformer}
An effective view transformation module is crucial for generating high-quality 3D features. 
Similar to FlashOcc \citep{yu2023flashocc}, we adopt Lift-Splat-Shoot (LSS) \citep{philion2020lift}, which predicts a per-pixel depth distribution and lifts image features using camera intrinsics and extrinsics. A hard first-surface target concentrates the depth distribution and its gradients at the LiDAR-observed surface, whereas semantic occupancy is supervised throughout a volume that includes structural thickness and occluded regions. The final DualPathOcc model therefore learns its lifting distribution from the occupancy objective without an explicit depth loss. Section~\ref{sec:gaussian_depth} compares one-hot, Gaussianized, and no-depth variants to study this surface--volume interaction.

\subsubsection{Spatial Enhancer}
The \textit{Spatial Enhancer} reduces information loss before BEV compression. While the FlashOcc configuration in our ablation aggregates lifted features on a \(200\times200\) grid, DualPathOcc retains a denser intermediate grid during early aggregation.

LSS first produces \(B\times C\times800\times800\times4\) features. Following the channel-to-height formulation, the four height bins are folded into channels to form \(B\times(4C)\times800\times800\). Two 2D convolution--downsampling stages then produce a \(B\times(4C)\times200\times200\) feature map for the dual-path encoder. The aggregate effect of this higher-resolution aggregation stage is measured in Section~\ref{sec:ablation}, and its computational trade-off is quantified in Section~\ref{sec:efficiency}.

\begin{figure}[!t]
  \centering{\includegraphics[width=0.6\linewidth]{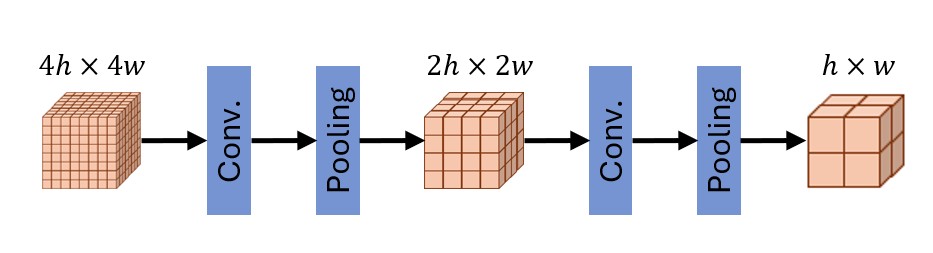}}\\
  \caption{Spatial Enhancer. After folding four height bins into channels, two 2D convolution--downsampling stages transform $B\times(4C)\times800\times800$ lifted features into the $B\times(4C)\times200\times200$ representation used by the dual-path BEV encoder.}
  \label{figure:spatial_enhancer}
\end{figure}

\subsubsection{Dual-path BEV Encoder}
A key challenge in camera-based occupancy prediction is obtaining dense and discriminative 3D features due to sparse voxelization when lifting 2D image features, as discussed in Sec.~\ref{sec:intro}. To address this, we propose a SENet-augmented dual-path BEV encoder.

The encoder adopts a dual-path design to balance fine-grained detail preservation and global context aggregation. It consists of a local path with high spatial resolution for detailed geometric cues and a global path with stronger downsampling to capture broader contextual information. The input BEV feature map \(F_{\text{in}}\) is processed through these two branches:

\begin{equation}
    F_{\text{local}} = \mathcal{E}_{\text{local}}(F_{\text{in}}) 
    \quad 
    F_{\text{global}} = \mathcal{E}_{\text{global}}(F_{\text{in}})
\end{equation}
and the outputs are subsequently fused to form a unified representation:
\begin{equation}
    F_{\text{BEV}} = \mathcal{F}(F_{\text{local}}, F_{\text{global}})
\end{equation}
where \(\mathcal{E}_{\text{local}}\) and \(\mathcal{E}_{\text{global}}\) denote the local and global encoders, and \(\mathcal{F}(\cdot)\) represents the local-global feature fusion module.

Each basic block incorporates SE channel recalibration \citep{hu2018squeeze}. Global average pooling summarizes channel responses, and a bottleneck MLP with sigmoid gating rescales the feature map. Applied after height is folded into channels, this operation emphasizes informative components of the sparse lifted BEV representation.

\subsubsection{Training Objective}
For supervision, we employ height-aware weighted cross-entropy that assigns larger weights to lower-elevation voxels, where occupied scene structure is concentrated. Let \(z_i\) denote the elevation bin of voxel \(i\), indexed upward from the lowest bin, and let \(w(z_i)\) denote its height-dependent weight. The objective is
\begin{equation}
    \mathcal{L}_{\text{occ}} = \frac{1}{N}\sum_{i=1}^{N}
    w(z_i)\,\ell_{\mathrm{CE}}(\mathbf{p}_i,y_i),
\end{equation}
where \(\mathbf{p}_i\) and \(y_i\) are the predicted class distribution and semantic label. We use a maximum lower-elevation weight of 3. This formulation changes the contribution of each voxel according to elevation while retaining the standard multiclass occupancy objective.

\section{Experiment}
\subsection{Datasets and Metrics}
\subsubsection{Datasets}
The occupancy benchmark is built on nuScenes \citep{caesar2020nuscenes,tian2023occ3d}. Voxel-wise annotations are provided within a 3D region of $[-40, -40, -1]$ to $[40, 40, 5.4]$ meters in the ego-centric coordinate system, with a voxel resolution of 0.4 m. The dataset includes 18 semantic classes, including one for free space. It also provides camera visibility masks indicating whether each voxel is observable from the available camera views.

\subsubsection{Metrics}
The models are mainly evaluated based on mIoU, which can be formulated as:
\begin{equation}
    mIoU=\frac{1}{C}\sum_{c=1}^C \frac{TP_c}{TP_c+FP_c+FN_c}
\end{equation}

where $TP_c$, $FP_c$, and $FN_c$ represent the number of true positive, false positive, and false negative predictions for class $c$, and $C$ is the total number of classes.

\subsection{Implementation Details}
\subsubsection{Training Strategies}
The proposed model was developed and trained on two different GPU workstations. 
During the development stage up to the \emph{Spatial Enhancer} module, training was conducted with a batch size of 4 on three NVIDIA RTX A6000 GPUs. 
For the full model, we used a batch size of 4 on eight NVIDIA H200 GPUs. 
In both cases, the AdamW optimizer was employed with a learning rate of \(1\times10^{-4}\), a weight decay of 0.05, and a total of 24 training epochs.

\subsubsection{Network Details}
The source camera images have resolution \(900\times1600\) and are processed at \(256\times704\). Standard data augmentation, including random horizontal flipping and rotation in image and 3D space, is applied during training. LSS produces \(C=64\) channels on an \(800\times800\) lateral grid with four height bins. Folding height into channels yields \(256\times800\times800\) features, and the Spatial Enhancer reduces them to \(256\times200\times200\) for 2D BEV processing.

The resulting BEV representation is fed into the dual-path encoder. The local path uses a downsampling ratio of (1/8), while the global path applies larger ratios of (1/16) and (1/32) to capture broader context. The fused features form a BEV map of size \(512 \times 200 \times 200\), which is passed to a lightweight 2D convolutional head for voxel-wise occupancy prediction.


\begin{table*}[!t]
\caption{3D occupancy prediction performance on the Occ3D-nuScenes validation set. ``Const. Veh." and ``Dri. Sur." denote construction vehicle and drivable surface, respectively.}

\resizebox{\textwidth}{!}
{
\begin{tabular}{l|lllllllllllllllll|l}

Method &
  \rotatebox{90}{\textcolor{undefined}{\rule{5pt}{5pt}}others} &
  \rotatebox{90}{\textcolor{barrier}{\rule{5pt}{5pt}}barriers} &
  \rotatebox{90}{\textcolor{bicycle}{\rule{5pt}{5pt}}bicycle} &
  \rotatebox{90}{\textcolor{bus}{\rule{5pt}{5pt}}bus} &
  \rotatebox{90}{\textcolor{car}{\rule{5pt}{5pt}}car} &
  \rotatebox{90}{\textcolor{consveh}{\rule{5pt}{5pt}}Const. veh.} &
  \rotatebox{90}{\textcolor{motorcycle}{\rule{5pt}{5pt}}motorcycle} &
  \rotatebox{90}{\textcolor{pedestrian}{\rule{5pt}{5pt}}pedestrian} &
  \rotatebox{90}{\textcolor{trafficcone}{\rule{5pt}{5pt}}traffic cone} &
  \rotatebox{90}{\textcolor{trailer}{\rule{5pt}{5pt}}trailer} &
  \rotatebox{90}{\textcolor{truck}{\rule{5pt}{5pt}}truck} &
  \rotatebox{90}{\textcolor{drivesurface}{\rule{5pt}{5pt}}Dri. Sur.} &
  \rotatebox{90}{\textcolor{otherlane}{\rule{5pt}{5pt}}other flat} &
  \rotatebox{90}{\textcolor{sidewalk}{\rule{5pt}{5pt}}sidewalk} &
  \rotatebox{90}{\textcolor{terrain}{\rule{5pt}{5pt}}terrain} &
  \rotatebox{90}{\textcolor{manmade}{\rule{5pt}{5pt}}manmade} &
  \rotatebox{90}{\textcolor{vegetation}{\rule{5pt}{5pt}}vegetation} &
  \rotatebox{90}{mIoU} \\ 
  
  \hline
  
MonoScene \cite{cao2022monoscene} & 1.7 & 7.2 & 4.2 & 4.9 & 9.3 & 5.6 & 3.9 & 3.0 & 5.9 & 4.4 & 7.1 & 14.9 & 6.3 & 7.9 & 7.4 & 1.0 & 7.6 & 6.0 \\
OccFormer \cite{zhang2023occformer} & 5.9 & 30.2 & 12.3 & 34.4 & 39.1 & 14.4 & 16.4 & 17.2 & 9.2 & 13.9 & 26.3 & 50.9 & 30.9 & 34.6 & 22.7 & 6.7 & 6.9 & 21.9 \\
TPVFormer \cite{huang2023tri} & 7.2 & 38.9 & 13.6 & 40.7 & 45.9 & 17.2 & 19.9 & 18.8 & 14.3 & 26.6 & 34.1 & 55.6 & 35.4 & 37.5 & 30.7 & 19.4 & 16.7 & 27.8\\
CTF-Occ \cite{tian2023occ3d} & 8.09 & 39.33 & 20.56 & 38.29 & 42.24 & 16.93 & 24.52 & 22.72 & 21.05 & 22.98 & 31.11 & 53.33 & 33.84 & 37.98 & 33.23 & 20.79 & 18.0 & 28.53 \\
RenderOcc \cite{pan2024renderocc} & 4.84 & 31.72 & 10.72 & 27.67 & 26.45 & 13.87 & 18.2 & 17.67 & 17.84 & 21.19 & 23.25 & 63.2 & 36.42 & 46.21 & 44.26 & 19.58 & 20.72 & 26.11 \\
PanoOcc \cite{wang2024panoocc} & 8.6 & 43.7 & 21.6 & 42.5 & 49.9 & 21.3 & \textbf{25.3 }& 22.9 & 20.1 & 29.7 & \textbf{37.1} & 80.9 & 40.3 & 49.6 & 52.8 & 39.8 & 35.8 & 36.6 \\
FlashOcc \cite{yu2023flashocc} &6.74&39.65&10.26&39.55&44.36&14.88&13.4&15.79&15.38&27.44&31.73&78.82&37.98&48.7&52.5&37.89&32.24&32.08   \\
DHD-S \cite{wu2025deep} & \textbf{10.59} & 43.21 & \textbf{23.02} & 40.61 & 47.31 & \textbf{21.68} & 23.25 & 23.85 & 23.40 & \textbf{31.75 }& 34.15 & 80.16 & 41.30 & 49.95 & 54.07 & 38.73 & 33.51 & 36.50 \\ 
ProtoOcc \cite{kim2025protoocc} & 11.09 & 43.37 & 22.64 & 42.25 & 49.71 & 19.33 & 24.83 & \textbf{26.08} & \textbf{25.56} & 27.71 & 34.77 & 79.79 & \textbf{43.56} & 50.89 & 54.01 & 40.04 & 34.03 & 37.04 \\
\hline
DualPathOcc &9.51&\textbf{47.01}&19.82& \textbf{42.54} &\textbf{50.32}&21.13&19.58&23.4&22.93&31.54& 34.81 &\textbf{81.84}&42.8&\textbf{52.42}&\textbf{56.51}&\textbf{42.91}&\textbf{36.15}&\textbf{37.37} \\ \hline
\end{tabular}
}
\label{tab:comparison_table}
\end{table*}

\begin{figure*}[!t]
  \centering{\includegraphics[width=0.8\linewidth]{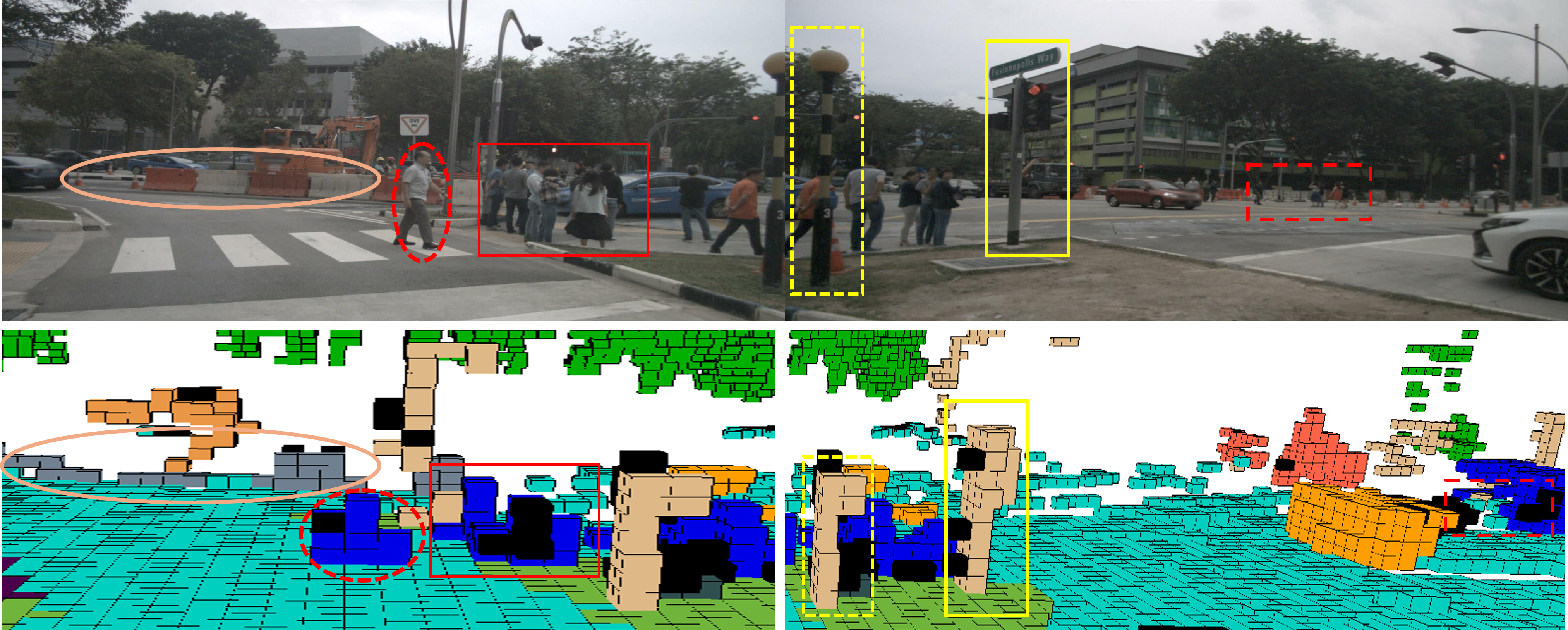}}\\
    \caption{
    Illustrative DualPathOcc predictions on Occ3D-nuScenes. Top: front and front-right camera views. Bottom: the corresponding semantic occupancy prediction. Voxels follow the class colors in Table~\ref{tab:comparison_table}, and outlines associate objects between image and occupancy views.
    }
  \label{figure:visualize_occ}
\end{figure*}

\subsection{Performance Comparison}

We compare DualPathOcc with representative published single-frame methods on the Occ3D-nuScenes validation set. Table~\ref{tab:comparison_table} summarizes the comparison.

DualPathOcc obtains 37.37 mIoU, 0.33 points above ProtoOcc (37.04), the strongest single-frame result in the table, and 5.29 points above FlashOcc (32.08). Relative to ProtoOcc, DualPathOcc is higher on 11 of 17 semantic categories, with gains of 3.83 for \textit{trailer}, 3.64 for \textit{barrier}, and 2.87 for \textit{manmade}, and lower on the remaining six: \textit{others}, \textit{bicycle}, \textit{motorcycle}, \textit{pedestrian}, \textit{traffic cone}, and \textit{other flat}. Figure~\ref{figure:visualize_occ} illustrates the semantic structure of representative predictions.

We also evaluate a temporal extension that aligns and fuses one historical frame with the current observation. At \(256\times704\), the \(800\times800\) variant reaches 39.72 mIoU. Table~\ref{tab:multi-frame} lists this lightweight extension alongside published multi-frame methods. It remains below the strongest of them, including ProtoOcc at 45.02, and measures what a single aligned frame contributes to the present architecture rather than serving as a competitive temporal result. Richer learned correspondence and temporal alignment remain promising directions.

\begin{table}[!t]
\caption{Component configurations on Occ3D-nuScenes. All variants share the $256\times704$ image input, batch size, AdamW optimizer, 24-epoch schedule, and no explicit depth loss. The variants are not capacity-matched because the listed modules add capacity; the Spatial Enhancer changes the intermediate grid from $200\times200$ to $800\times800$ before returning to $200\times200$.}
\centering
\begin{tabular}{lc}
\hline
  Configuration & mIoU \\ \hline
FlashOcc baseline & 32.08 \\
Baseline + Spatial Enhancer & 35.31 \\
Baseline + height-aware loss & 35.71 \\
Baseline + dual-path BEV encoder & 36.20 \\
Baseline + dual-path BEV encoder + SENet & 36.92\\
Encoder + SENet + Spatial Enhancer & 37.15\\
Encoder + SENet + Spatial Enhancer + height-aware loss & 37.37\\
\hline
\end{tabular}
\label{tab:dual_channel_ablation}
\end{table}

\begin{table}[!t]
\caption{Comparison with multi-frame methods on the Occ3D-nuScenes validation set. 
DualPathOcc variants use one aligned historical frame. The base model uses a $400\times400$ Spatial Enhancer grid and batch size 4; * uses $800\times800$ and batch size 2. The \textasciitilde{} and \textasciicircum{} variants use $512\times1408$ images with $400\times400$ and $800\times800$ grids, respectively, and batch size 2.
}
\centering
\begin{tabular}{cccc}
\hline
Model & Image Backbone & Image Size & mIoU   \\ \hline
BEVFormer & ResNet-101 & $928\times1600$ & 26.88 \\
FastOcc & ResNet-101 & $928\times1600$ & 39.21 \\
PanoOcc & ResNet-101 & $928\times1600$ & 42.13 \\
BEVDet4D & ResNet-50 & $384\times704$ & 39.25\\
FB-OCC & ResNet-50 & $256\times704$ & 40.69\\
FlashOcc & ResNet-50 & $256\times704$ & 37.84 \\
COTR & ResNet-50 & $256\times704$ & 44.45\\
ProtoOcc & ResNet-50 & $256\times704$ & 45.02\\
\hline
DualPathOcc & ResNet-50 & $256\times704$ & 38.69\\
DualPathOcc* & ResNet-50 & $256\times704$ & 39.72\\
DualPathOcc\textsuperscript{\textasciitilde} & ResNet-50 & $512\times1408$ & 40.7\\
DualPathOcc\textsuperscript{\textasciicircum} & ResNet-50 & $512\times1408$ & 41.9\\
\hline
\end{tabular}
\label{tab:multi-frame}
\end{table}

\subsection{Computational Profile}
\label{sec:efficiency}
We profile the trained DualPathOcc and FlashOcc checkpoints under the same inference protocol on one RTX A6000: batch size 1, FP32, synchronized timing without data loading, 50 warm-up iterations, and three runs of 500 timed iterations. DualPathOcc/FlashOcc require 224.6/103.0 ms (4.45/9.71 FPS), 370.9/44.7M parameters, and 4048.7/728.9 MiB peak allocated memory. MMCV counts 2.094/0.249 TFLOPs over supported operators; custom CUDA and functional operations are excluded. Relative to FlashOcc, DualPathOcc therefore requires 2.18$\times$ the latency, 8.29$\times$ the parameters, and 5.55$\times$ the peak memory for a 5.29-point aggregate mIoU difference. ProtoOcc reports 77.9 ms, or 12.83 FPS, for its single-frame ResNet-50 model at $256\times704$ on an RTX 3090 \citep{kim2025protoocc}, measured on different hardware from the protocol used here.

\subsection{Ablation Study}
\label{sec:ablation}

Table~\ref{tab:dual_channel_ablation} reports the available component configurations on Occ3D-nuScenes. The 32.08 baseline is the single-frame FlashOcc configuration: LSS features are folded from $B\times C\times H\times W\times Z$ to $B\times(CZ)\times H\times W$ and processed with 2D BEV convolutions on a $200\times200$ grid. The variants share the image input and training recipe, while their capacity follows the listed components.

\subsubsection{Spatial Enhancer}

Adding the \textit{Spatial Enhancer} to the baseline increases mIoU from 32.08 to 35.31, a 3.23-point gain. The measurement is an aggregate over all semantic classes from a single training run per configuration; these experiments do not resolve which classes or spatial scales account for the difference.

\subsubsection{Height-aware Loss}

The height-aware objective obtains 35.71 mIoU when added to the baseline. By assigning greater weight to lower-elevation voxels, it adapts the occupancy objective to the vertical distribution of driving scenes.

\subsubsection{SENet-augmented Dual-path BEV Encoder}
The dual-path BEV encoder obtains 36.20 mIoU, and adding SENet channel recalibration increases the result to 36.92. Combining the encoder, SENet, and Spatial Enhancer reaches 37.15; the complete configuration with height-aware weighting reaches 37.37. Together, these configurations show complementary gains from high-resolution aggregation, local--global encoding, and elevation-aware optimization.

\subsubsection{
Depth Supervision}
We investigate the effect of depth supervision on image-based 3D occupancy prediction. As shown in Table~\ref{tab:sigma_comparison}, training with one-hot depth supervision yields an mIoU of 35.96, which is nearly one point lower than the model trained without depth supervision, suggesting a mismatch between surface-based depth signals and volumetric occupancy learning.

The no-depth variant obtains 36.92 mIoU. Replacing the one-hot target with a Gaussian distribution centered at the measured depth ($\sigma=1.5$) gives 36.94 mIoU, a comparable result. The contrast with one-hot supervision indicates that relaxing the sharp surface target avoids its degradation, while the final 37.37 model uses no explicit depth loss. Implementation details are provided in the appendix.

\begin{table}[!t]
\centering
\caption{Comparison of models trained with different depth supervision strategies on the Occ3D-nuScenes dataset. All models in this experiment exclude the Spatial Enhancer module.}
\begin{tabular}{cc}
\hline
Depth supervision method  & mIoU \\ 
\hline
One-hot & 35.96 \\ 
No supervision & 36.92 \\ 
Gaussian ($\sigma=1.5$) & 36.94 \\
\hline
\end{tabular}
\label{tab:sigma_comparison}
\end{table}
\color{black}

\section{Limitations}
DualPathOcc offers several directions for further development. The temporal extension uses one historical frame with simple alignment; learned correspondence over longer histories could provide richer motion and occlusion cues. The $800\times800$ intermediate representation increases latency, parameter count, and memory for a 3.23-point aggregate mIoU difference, motivating sparse or adaptive-resolution aggregation. The 37.37 result is the best checkpoint from one training run, so multi-seed evaluation would characterize its variance. Finally, experiments are currently limited to Occ3D-nuScenes, and cross-dataset evaluation will be valuable for measuring generalization across sensing conditions and scene distributions.

\section{Conclusion}
\label{sec: conclusion}
We present DualPathOcc, a camera-based 3D occupancy framework that combines high-resolution aggregation before BEV compression, a SENet-augmented dual-path encoder, and height-aware occupancy weighting. DualPathOcc achieves 37.37 mIoU on single-frame Occ3D-nuScenes, 0.33 points above ProtoOcc. The depth analysis further shows that Gaussianized and no-depth training provide comparable occupancy accuracy, while a hard surface-centered target is less aligned with volumetric prediction.

\bibliography{example}  

\newpage
\appendix
\section{Depth Supervision}
\label{sec:gaussian_depth}
\begin{figure}[!ht]
  \centering{\includegraphics[width=0.8\linewidth]{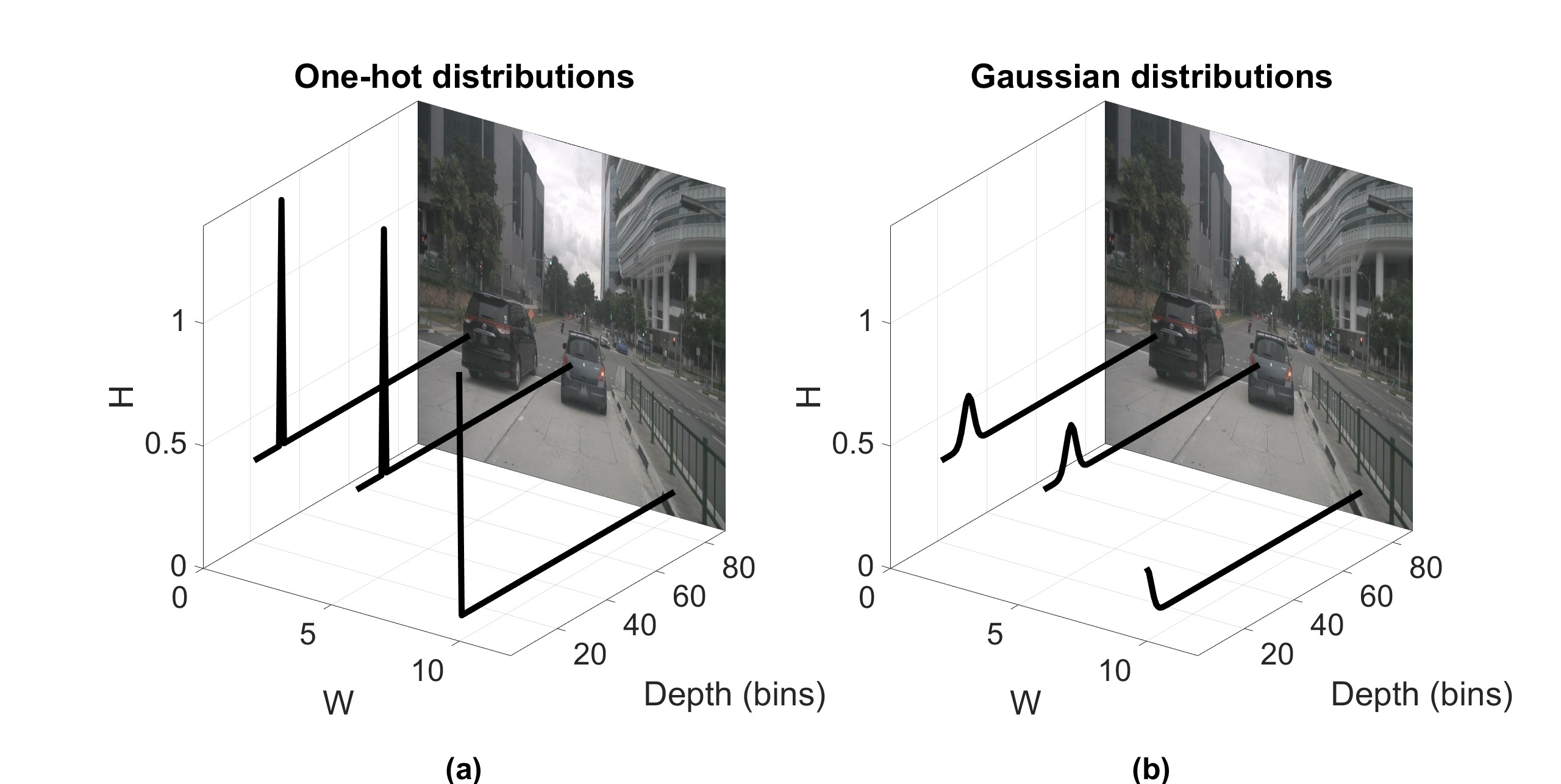
  }}\\
  \caption{Comparison of ground-truth depth representations on a sample image. 
(a) One-hot encoded depth distribution. 
(b) Gaussian-encoded depth distribution. 
$H$ and $W$ denote image height and width. 
Each depth bin corresponds to the distance of a pixel projected into 3D space relative to the ego vehicle. 
In the one-hot encoding, each pixel activates only a single bin with a value of 1, while in the Gaussian encoding, the area under the Gaussian curve for each pixel sums to 1, providing a smoother depth representation.}
\label{figure:depth_dist}
\end{figure}

The final DualPathOcc model uses no explicit depth loss. This appendix describes the one-hot and Gaussianized targets used in the diagnostic comparison of Table~\ref{tab:sigma_comparison}. In prior work~\cite{wu2025deep, zhang2023occformer}, depth supervision is commonly imposed on a one-hot depth distribution obtained by projecting LiDAR points into the image plane. The discrete one-hot target at pixel \(i\) is
\begin{equation}
y_{i,d} =
\begin{cases}
1, & \text{if } d = \left\lfloor \dfrac{D_i}{\Delta d} \right\rfloor, \\
0, & \text{otherwise},
\end{cases}
\label{eq:one_hot_depth}
\end{equation}
where \(D_i\) is the LiDAR-projected depth at pixel \(i\), \(\Delta d\) is the discretized depth-bin width, \(d\) indexes depth bins, and \(y_{i,d}\) is the one-hot depth label (see Figure~\ref{figure:depth_dist}(a)). While this target enforces a precise first-surface depth, it yields an excessively peaked supervision signal that concentrates all probability mass on the first visible surface along a ray. Such sharp supervision can be misaligned with the volumetric nature of 3D occupancy prediction and may degrade occupancy performance.

For the Gaussianized diagnostic, we soften the one-hot target by applying a Gaussian kernel along the depth dimension. The unnormalized response centered at the LiDAR depth \(D_i\) is
\begin{equation}
\tilde{y}_{i,d} \;=\; \exp\!\left(-\,\frac{\big(d\Delta d - D_i\big)^2}{2\sigma^2}\right),
\label{eq:gauss_unnorm}
\end{equation}
and the normalized Gaussianized depth target is
\begin{equation}
y_{i,d} \;=\; \frac{\tilde{y}_{i,d}}{\sum_{k=0}^{N_d-1}\tilde{y}_{i,k}} .
\label{eq:gauss_norm}
\end{equation}
Here \(\sigma\) controls the spread of the depth likelihood, and \(N_d\) denotes the number of discretized depth bins. The normalized vector \(y_{i,\cdot}\) encodes a soft likelihood across neighboring bins rather than a single spike (Figure~\ref{figure:depth_dist}(b)). With \(\sigma=1.5\), this target obtains 36.94 mIoU, comparable to 36.92 without depth supervision and above the 35.96 one-hot result.

For this diagnostic variant, the predicted per-pixel distribution \(\hat{p}_{i,d}\) is trained with soft cross-entropy:
\begin{equation}
\mathcal{L}_{\text{depth}} \;=\; -\frac{1}{N_{\text{pix}}}\sum_{i}\sum_{d} y_{i,d}\,\log \hat{p}_{i,d},
\label{eq:depth_ce}
\end{equation}

\subsubsection{Qualitative and Discussion Analysis}
\begin{figure*}[!ht]
    \centering

    \subfloat[]{\includegraphics[width=0.45\linewidth]{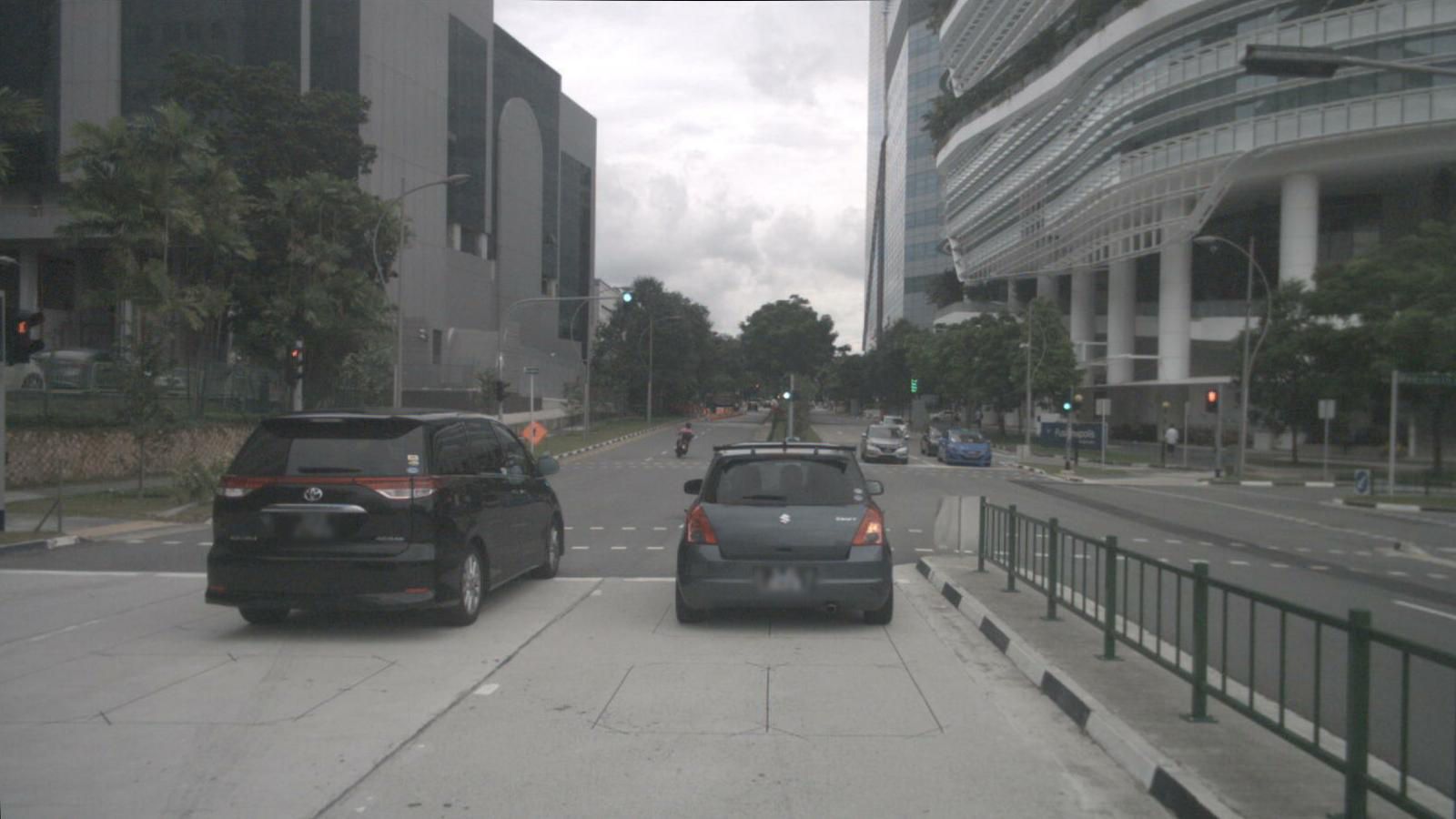}\label{fig:first}}
    \hfill
    \subfloat[]{\includegraphics[width=0.45\linewidth]{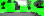}\label{fig:second}}

    \vspace{2mm}

    \subfloat[]{\includegraphics[width=0.45\linewidth]{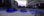}\label{fig:third}}
    \hfill
    \subfloat[]{\includegraphics[width=0.45\linewidth]{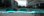}\label{fig:fourth}}

    \caption{Qualitative comparison of depth supervision strategies. 
    (a) Original camera input showing two nearby vehicles. 
    (b) One-hot depth supervision, where green pixels indicate voxels classified at the 7\,m distance bin. 
    (c) Model trained without depth supervision, predicting the 7\,m region with weaker spatial focus (blue). 
    (d) Gaussian-encoded depth supervision, showing an intermediate spread of responses in the 7\,m bin (cyan).}
    \label{fig:depth_visualization}
\end{figure*}

\begin{figure*}[!ht]
    \centering
    \subfloat[]{%
        \includegraphics[width=0.32\textwidth]{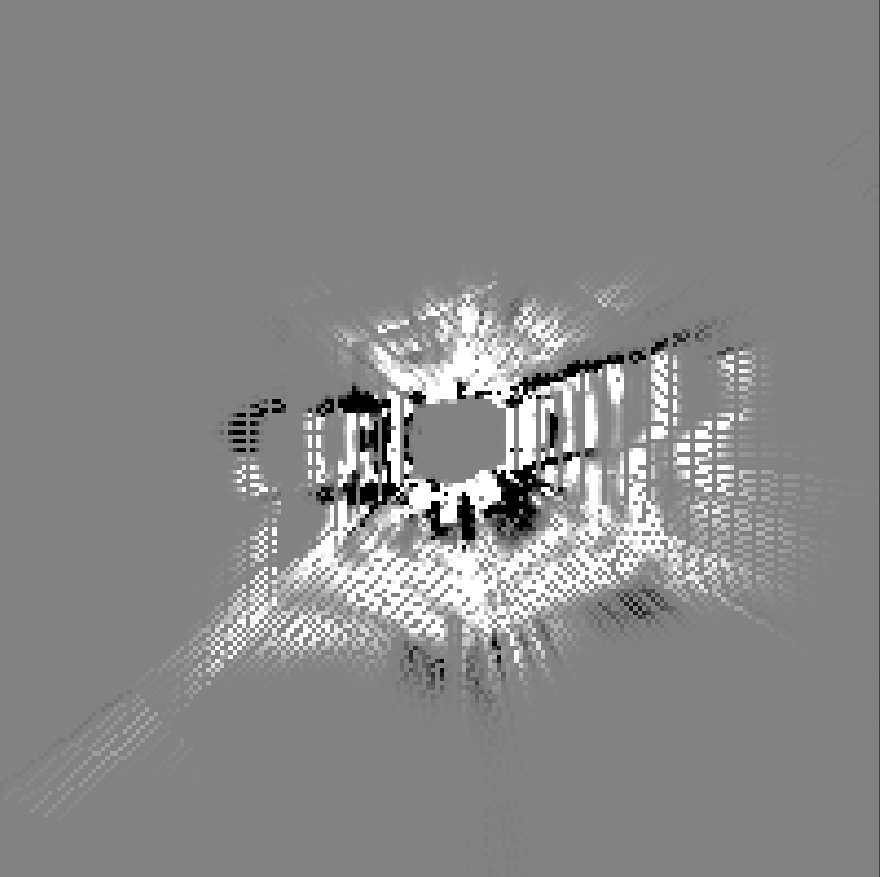}%
        \label{fig:onehot}}
    \hfill
    \subfloat[]{%
        \includegraphics[width=0.32\textwidth]{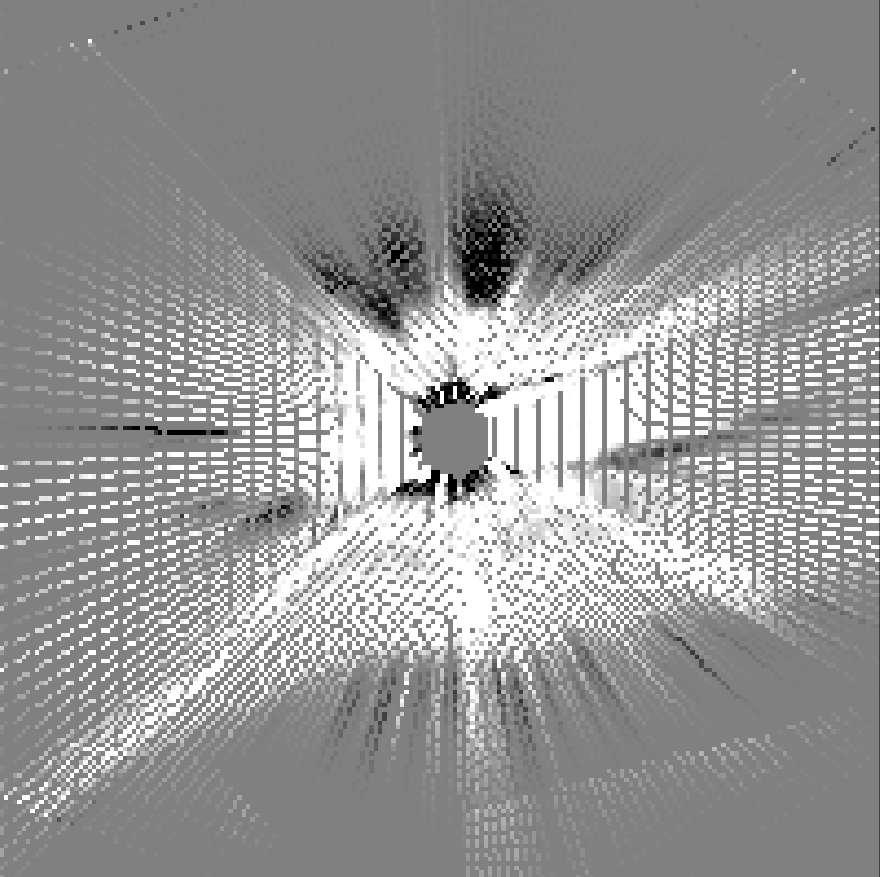}%
        \label{fig:no_loss}}
    \hfill
    \subfloat[]{%
        \includegraphics[width=0.32\textwidth]{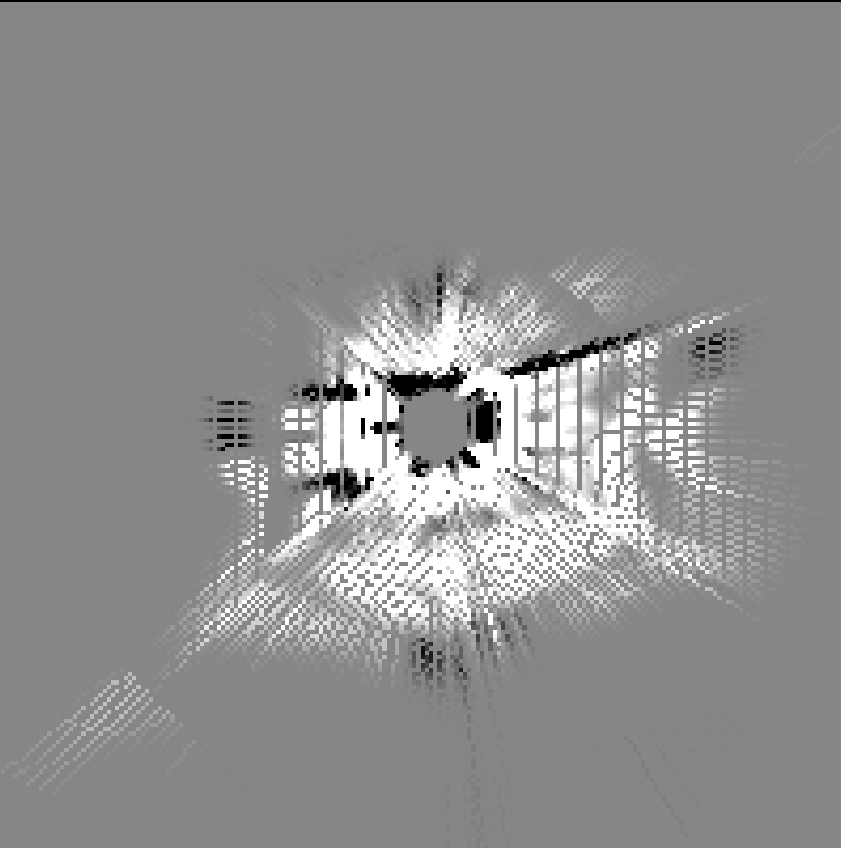}%
        \label{fig:gaussian}}
    \label{fig:lss_depth_variants}
    \caption{Comparison of BEV feature projections under different depth supervision strategies. 
(a) One-hot encoded depth supervision, 
(b) no depth supervision, and 
(c) Gaussian-encoded depth supervision. 
Gray regions indicate undefined areas, while white and black denote high and low feature responses, respectively.}
    \label{fig:lss_feats}
\end{figure*}

We visualize predicted depth distributions and corresponding same-channel BEV features after LSS (Figures~\ref{fig:depth_visualization} and~\ref{fig:lss_feats}). These examples illustrate how the supervision target changes feature concentration along image rays and across the BEV plane.

The one-hot depth supervision yields the sharpest and most localized depth features, accurately regressing object distances. However, its corresponding BEV representation is spatially limited, with strong activations concentrated near the ego vehicle.

In contrast, the model trained without depth supervision produces noticeably less accurate depth maps---its per-pixel depth confidence is diffuse and inconsistent. Nevertheless, its BEV feature map exhibits a much broader spatial coverage, spreading across the entire $H$--$W$ plane. This suggests that the network, unconstrained by explicit depth regression, learns to encode broader spatial cues that benefit overall BEV reasoning despite weaker pixel-level accuracy.

Gaussianized supervision exhibits an intermediate behavior. Its depth responses are more localized than in the no-depth case but less concentrated than with one-hot encoding. Correspondingly, its BEV activations extend farther than those of the one-hot variant but are less widespread than those from the no-depth variant, consistent with the comparable aggregate results in Table~\ref{tab:sigma_comparison}.

\end{document}